\documentclass[conference]{IEEEtran}
\IEEEoverridecommandlockouts
\usepackage{cite}
\usepackage{subcaption}
\usepackage{amsmath,amssymb,amsfonts}
\usepackage{algorithmic}
\usepackage{graphicx}
\usepackage{textcomp}
\usepackage{xcolor}

\usepackage{listings}

\lstdefinestyle{mystyle}{
    basicstyle=\ttfamily\footnotesize,
    breakatwhitespace=false,         
    breaklines=true,                 
    captionpos=b,                    
    keepspaces=true,                 
    numbersep=5pt,                  
    showspaces=false,                
    showstringspaces=false,
    showtabs=false,                  
    tabsize=2,
    language=C++
}

\def\BibTeX{{\rm B\kern-.05em{\sc i\kern-.025em b}\kern-.08em
    T\kern-.1667em\lower.7ex\hbox{E}\kern-.125emX}}
\begin{document}

\title{Image Classification on Accelerated Neural Networks}

\author{
    \IEEEauthorblockN{İlkay Sıkdokur$^1$, İnci M. Baytaş$^2$, Arda Yurdakul$^{1,2}$}
    \IEEEauthorblockA{
    $^1$Department of Computational Science and Engineering, Bogazici University, Turkey}
    \IEEEauthorblockA{
    $^2$Department of Computer Engineering, Bogazici University, Turkey
    \\\{ilkay.sikdokur, inci.baytas, yurdakul\}@boun.edu.tr
    }
}

\maketitle

\begin{abstract}
For image classification problems, various neural network models are commonly used due to their success in yielding high accuracies. Convolutional Neural Network (CNN) is one of the most frequently used deep learning methods for image classification applications. It may produce extraordinarily accurate results with regard to its complexity. However, the more complex the model is the longer it takes to train. In this paper, an acceleration design that uses the power of FPGA is given for a basic CNN model which consists of one convolutional layer and one fully connected layer for the training phase of the fully connected layer. Nonetheless, inference phase is also accelerated automatically due to the fact that training phase includes inference. In this design, the convolutional layer is calculated by the host computer and the fully connected layer is calculated by an FPGA board. It should be noted that the training of convolutional layer is not taken into account in this design and is left for future research. The results are quite encouraging as this FPGA design tops the performance of some of the state-of-the-art deep learning platforms such as Tensorflow on the host computer approximately 2 times in both training and inference.
\end{abstract}

\section{Introduction}

\begin{figure*}[htbp]
\centering
\includegraphics[width=180mm,scale=0.5]{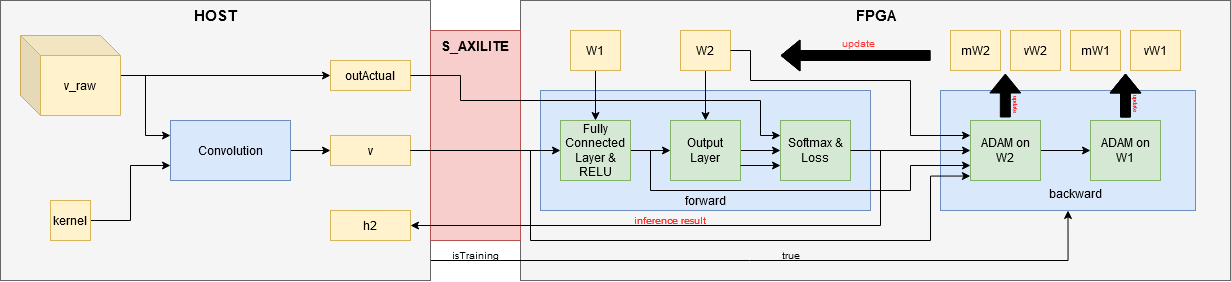}
\caption{Sketch of the design}
\label{fig}
\end{figure*}

Image classification is one of the core tasks in many artificial intelligence tasks, such as autonomous driving, object recognition, and robotics applications. Such applications require real-time image classification where the latency and the size of the model become some of the important factors that define the performance of the task. Convolutional Neural Network (CNN), a deep learning architecture for grid-like inputs (e.g., image, graph), constitutes the state-of-the-art in image classification. Although CNN is successfully used for image classification, high model complexity of CNN training leads to high computational time and memory consumption. For this reason, CNN architectures may be cumbersome when it comes to using them in real-time applications. This situation rises the importance of acceleration of CNN models that are used in image classification problems.

For the acceleration of deep neural networks, different types of hardware units are used such as Graphic Processing Unit (GPU) and Field-Programmable Gate Array (FPGA). FPGA makes an outstanding candidate for this acceleration process in various ways. One of the reasons of this is that FPGA is a fully configurable system which consists of numerous fully programmable logic gates. This feature of FPGA makes it easier and cheaper to specialize by the designer for the acceleration of any desired calculation. However, the programmable logic gates need to be programmed via a low level language such as Hardware Description Language (HDL) in general like the other hardware units. But this problem can be solved by the second advantage of FPGA usage: High Level Synthesis (HLS). HLS method lets the designer to write the code piece that is sought to be accelerated in various high level languages such as C++, transforms it into low level hardware configuration languages such as HDL and simulates a numerical circuit as if the logic gates are programmed to do the desired calculations as stated in the high level. Besides, it lets the designer to be more flexible in the design of the circuit for acceleration by giving options to the designer on how the acceleration is sought to be done like the parallelization methods and factors, memory usage on the unit, etc. It is much more easier to change the acceleration design in numerous aspects of it compared to other hardware units. By considering all of these reasons, the acceleration of CNN training is decided to be done on an FPGA board in this study. The target FPGA board is chosen as FFV1156 package of Zynq-7000 board family in this study. Moreover, the host computer that the FPGA is connected to runs a 64 bit operating system and contains a CPU as AMD Ryzen 5 3550H and 8 GB DDR4 RAM. 

In this study, the goal is to offer a hardware design that is to accelerate the training phase of a basic CNN model that has one convolutional layer, one fully connected layer and one output layer as shown in Fig. 1. Since the inference phase forward propagation is contained in the training phase forward and backward propagation, this design accelerates the inference phase also. The results of acceleration of inference phase is also given in the paper with the acceleration of the training phase. It should be noted that in this study, training includes learning the parameters of the fully connected layers. The convolution kernels are not learned from data. However, the convolution operation is still performed in the forward propagation such that the complexity of the convolution is included in the inference phase. Training of the convolution layer is left for future work. In this sense, convolutional kernel is set to be a constant array that is $3\times3$ sharpening kernel being commonly used in image processing applications. The kernel can be seen in Eq. (1):

\begin{equation}
kernel=\begin{bmatrix}
0 & -1 & 0\\ 
-1 & 5 & -1\\ 
0 & -1 & 0
\end{bmatrix}
\end{equation}

The kernel is illustrated with \textbf{kernel} in the Fig. 1. The convolution operation is calculated in the host computer whereas the fully connected layer and the output layer is calculated in the FPGA board.

To describe the model and the dataset, the fully connected layer consists of 128 neurons. The input images used in this problem are 28x28 pixel MNIST handwritten digits (LeCun et al., 1998). Therefore, the output layer is of 10 neurons for each digit from 0 to 9. It should be noted that mini-batch normalization is used that is of 32 images. In Fig. 1, these mini-batches are illustrated with \textbf{v\_raw} and the classes for every image in the mini-batches are illustrated with \textbf{outActual}. Firstly, a 2D convolution is applied on each mini-batch by the 3x3 kernel with stride of 1 and padding of 0 in both directions and then their sized are reduced into half by 2x2 max-pooling. The convolved mini-batches are flattened and inserted into the model. These flattened images are illustrated with \textbf{v} in Fig. 1. The outputs obtained after these flattened images are inserted into the model is illustrated as \textbf{h2} in Fig. 1. MNIST dataset consists of 60,000 images for training and 10,000 images for test purpose (LeCun et al., 1998). For training, both forward and backward passes are done for 60,000 training images and for testing, only forward pass is done for 10,000 testing images. Adam optimizer (Kingma \& Ba, 2014) is used during training phase. Firstly, step-wise moving momentums ($m_t$) are calculated by using the momentum values of previous step ($m_{t-1}$) and the gradient of the loss of the current mini-batch ($g_t$) as in Eq. (2):

\begin{equation}
\begin{aligned}
m_t = \beta _1m_{t-1}+(1-\beta _1)g_t\\
v_t = \beta _2v_{t-1}+(1-\beta _2)g_t^2
\end{aligned}
\end{equation}

And then, in order to make momentums move and converge slower as the steps increase, a step-wise correction is applied on the momentums as in Eq. (3):

\begin{equation}
\begin{aligned}
\widehat{m_t} = \frac{m_t}{1-\beta_1 ^t}\\
\widehat{v_t} = \frac{v_t}{1-\beta_2^t}
\end{aligned}
\end{equation}

These corrected momentums are illustrated as \textbf{mW1}, \textbf{vW1}, \textbf{mW2} and \textbf{vW2} in Fig. 1. Finally, the weights are updated by using the updated and correction applied momentum values as in Eq. (4):

\begin{equation}
W_t = W_{t-1}-\eta \frac{\widehat{m}_t}{\sqrt{\widehat{v_t}}+\epsilon }
\end{equation}

In Fig. 1, the weights are illustrated as \textbf{W1} and \textbf{W2}.

\begin{table}[htbp]
\centering
\caption{ADAM constant values}
\begin{tabular}{|c|c|}
\hline
Constant                & Value                  \\ \hline
$\beta_1$ & $0.9$                  \\ \hline
$\beta_2$ & $0.999$                  \\ \hline
$\eta$     & $0.01$                  \\ \hline
$\epsilon$ & $10^{-7}$ \\ \hline
\end{tabular}
\end{table}

Hyperparameters of Adam optimizer are chosen in agreement with the literature as given in Table 1. Nonetheless, being one of the common practices, the initialization of the weights are done by Gaussian noise where the mean is 0 and the standard deviation is 0.1.

For the acceleration operations, matrix multiplication and addition loops that are mainly used during training are unrolled in order to grant parallelization by a specific factor. In order to grant parallelization, the arrays to be multiplied and added need to be available for parallel access. In order to create parallel accesses to the arrays, they are partitioned in cyclic manner by the same factor. Moreover, the calculations that are already done are stored in the memory in order to prevent a re-calculation. 

During the progress of this paper, a literature review of related works will take place in Section 2. In Section 3, the design will be described. In this description, it will be detailed how the design is structured for the model, how the parallelization is supplied and how the memory storage is handled in order to maintain this parallelization. Finally, final thoughts and some future work insights will be given in Section 4.

\section{Literature Review}
Current works in hardware acceleration of neural networks offer various techniques and designs of implementations of trained models on FPGA. Chitty-Venkata and Somani (2020) offer a method to reduce the size of the weight matrices into systolic arrays by pruning various nodes of weight arrays so they achieve a faster inference in less cycles. Arora, Wei and John (2020) propose adding additional matrix multiplier blocks of systolic arrays on FPGA in order to accelerate the process of FPGAs on machine learning applications. They obtain almost 4 times faster results than reference design. In another paper, Zhang, Zheng and Prasanna (2020) propose a design that aims to increase throughput in Graph Convolutional Networks that are based on data traffic of web sites. They part the data by blocks and cut unnecessary parts off so they achieve faster results. Sestito et al. (2020) offer a new method of convolution by applying a sliding window FIFO structure where multiple convolution layers work in parallel. They state that this module that they offer works approximately 650 times faster with 70 times less memory usage and 13 times less energy consumption. Meng et al. (2020) offer packing data and using ensemble models on the packages of data. They state that they obtain a speed-up between 2 and 7 times and there could even be a possible increase in accuracy due to usage of ensemble methods. Moreover, Gibson et al. (2020) offer a method in which they divide the kernel and images into parts, apply convolution separately on each part and consolidate the results afterwards. They state that they obtain up to 4 times speed-up than reference designs. Ting et al. (2020) offer a method to reschedule the works of multiple Deep Neural Network modules in multiple accelerators by deciding how accurate results they are sought to produce. By rescheduling the modules in an optimal way, they vanish the waiting periods between each module process so accelerate the whole process in overall. Dias et al. (2020) offer to process each dimension of multi-dimensional data in parallel and sum up the results afterwards in order to increase throughput. 

Results of some current papers show that FPGAs are very good candidates for implementation of the training of neural networks. Courbariaux et al. (2016) use Binarized Neural Networks to speed-up training and inference in which weight matrices consist of only -1 and 1 values. In addition to binarized values, Blott et al. (2018) propose usage of Quantized Neural Networks in which weights are limited in an interval which gives a better accuracy on FPGA. Prost-Boucle, Bourge and Pétrot (2018) offer usage of -1, 0 and 1 values for weights on FGPA in which they obtained remarkable speed-ups.

\section{Design}
\subsection{Description}
The sketch of the design can be seen in Fig. 1. In the figure, input and output arrays, different modules of the CNN model and the memory interface connection between the host and the FPGA is shown.
The size mappings for this problem can be seen in Table 2. The definitions and sizes of the arrays illustrated in the figures are given in Table 3. The system is available to use any dataset so the sizes can be changed freely.

\begin{table}[htbp]
\centering
\caption{Definitions and values of the model constants}
\begin{tabular}{|c|c|c|}
\hline
Constant                                                          & Definition                                                                                & Value                                           \\ 
\hline
BATCHSIZE                                                           & \begin{tabular}[c]{@{}c@{}}Image amount \\ in a mini-batch\end{tabular}                  & 32                                               \\
\hline
\begin{tabular}[c]{@{}c@{}}IMAGEX\\ IMAGEY\end{tabular}           & \begin{tabular}[c]{@{}c@{}}Width and height\\ of input image\end{tabular}                 & \begin{tabular}[c]{@{}c@{}}28\\ 28\end{tabular} \\ \hline
\begin{tabular}[c]{@{}c@{}}KERNELX\\ KERNELY\end{tabular}         & \begin{tabular}[c]{@{}c@{}}Width and height \\ of the kernel\end{tabular}                 & \begin{tabular}[c]{@{}c@{}}3\\ 3\end{tabular}   \\  \hline
POOLMAPLENGTH                                                     & \begin{tabular}[c]{@{}c@{}}Size of the flattened \\ product in the end\end{tabular}       & 169                                             \\ \hline
LAYERSIZE                                                         & \begin{tabular}[c]{@{}c@{}}Size of the \\ hidden layer\end{tabular}                       & 128                                             \\ \hline
CLASSSIZE                                                         & \begin{tabular}[c]{@{}c@{}}Amount of possible \\ classes in the dataset\end{tabular}      & 10                                              \\ \hline
\end{tabular}
\end{table}

\begin{table}[htbp]
\centering
\caption{Variables, definitions and sizes}
\begin{tabular}{|c|c|c|}
\hline
Array                                             & Definition                                                                                 & Size                                                                  \\ \hline
v\_raw                                            & Images from dataset                                                                        & \begin{tabular}[c]{@{}c@{}}BATCHSIZE$\times$\\ IMAGEX$\times$\\ IMAGEY\end{tabular} \\ \hline
kernel                                            & Convolution kernel                                                                         & \begin{tabular}[c]{@{}c@{}}KERNELX$\times$\\ KERNELY\end{tabular}            \\ \hline
outActual                                         & Classes of images                                                                          & \begin{tabular}[c]{@{}c@{}}BATCHSIZE$\times$\\ CLASSSIZE\end{tabular}        \\ \hline
v                                                 & Convolution product                                                                        & \begin{tabular}[c]{@{}c@{}}BATCHSIZE$\times$\\ POOLMAPLENGTH\end{tabular}    \\ \hline
W1                                                & \begin{tabular}[c]{@{}c@{}}Weights of fully \\ connected layer\end{tabular}                & \begin{tabular}[c]{@{}c@{}}POOLMAPLENGTH$\times$\\LAYERSIZE\end{tabular}    \\ \hline
W2                                                & Weights of output layer                                                                    & \begin{tabular}[c]{@{}c@{}}LAYERSIZE$\times$\\ CLASSSIZE\end{tabular}        \\ \hline

h2                                                & \begin{tabular}[c]{@{}c@{}}Output layer values\\  after activation\end{tabular}            & \begin{tabular}[c]{@{}c@{}}BATCHSIZE$\times$\\ CLASSSIZE\end{tabular}        \\ \hline

\begin{tabular}[c]{@{}c@{}}mW1\\ vW1\end{tabular} & \begin{tabular}[c]{@{}c@{}}ADAM momentums of fully\\  connected layer weights\end{tabular} & \begin{tabular}[c]{@{}c@{}}POOLMAPLENGTH$\times$\\ LAYERSIZE\end{tabular}    \\ \hline
\begin{tabular}[c]{@{}c@{}}mW2\\ vW2\end{tabular} & \begin{tabular}[c]{@{}c@{}}ADAM momentums of \\ output layer weights\end{tabular}          & \begin{tabular}[c]{@{}c@{}}LAYERSIZE$\times$\\ CLASSSIZE\end{tabular}        \\ \hline
\end{tabular}
\end{table}

It can be seen from Fig. 1 that calculations commence on host side as convolution is applied on raw images taken from the dataset. Then, for the sake of application of supervised learning, the actual class vectors are prepared in order to calculate loss between them and obtained output values of the model. The output of the model is kept in the host side in order to be used in testing phase to test how accurate the model predicts the classes since this design allows an acceleration in inference as well. This is the flow of the host side in brief. In addition, the both of the weights are instantiated in host part.

It is seen that host part and FPGA part are connected with an S\_AXILITE interface. This part allows the usage of AXI-4 LITE inter-chip communication interface. The advantage of using AXI-4 interfaces is that they build a connection between the different parts of the design by creating registers between the parts. One of the main advantages of AXI-4 interfaces is that they also provide data burst features, i.e more than one pack of data could be read of written through the interface which provides a data stream. However, AXI-4 LITE interface provides only one pack is transferred from one part to the other. The reason of using this limited interface is that the other interfaces stream the data from one part to the other to be stored in an extra storage in order to be used more than once for the sake of keeping the data stream continuous. In this model, for example the products of convolution and output layer values are used more than once. By keeping aligned with the target FPGA choice, it is aimed to reduce storage usage as well in order to fit into the chip. Nonetheless, in future work the other interfaces could be used with a bigger and more complex FPGA chip to increase throughput of data even more.

It can be seen in Fig. 1 that the host sends a signal illustrated as $isTraining$ in the figure to the FPGA. This signal distinguishes whether the process is desired to be ended after the forward pass that is inference phase or kept going for whole training phase including backward pass. This design grants that the inference phase is accelerated as well since the whole training is aimed to be accelerated. 

Also, it can be noticed that almost every part of the modules in the FPGA use some values in its calculations that are calculated in the part before. This is a feature of the simplicity of this design in the FPGA part also. Unnecessary calculations are sought to be avoided so this decreases the amount of computations a lot. For example, RELU activation operations are calculated right in the end of fully connected layer calculations. Another example is that the sum of all exponentials of output layer values is calculated concurrently in output layer calculations for softmax activation operations. In the softmax module, cross-entropy loss is calculated in parallel. These are the main calculations done in forward pass module.

The same less computation approach could be seen in backward pass also. Since some of the values that are needed for the ADAM calculation on the fully connected layer weights are calculated before in ADAM calculation on the output layer weights, they are passed through one side through the other. The step-wise correction values consist of power operation which is a computationally expensive operation. In order to reduce the application of this operation, they are calculated once and used in the next ADAM calculation because they are based on the index of mini-batches that are passed through the model and both ADAM calculations are done for the same mini-batch. It can be seen that momentum values are seen to be updated. This is because ADAM is an adaptive optimization method and this leads the momentums are to be stored in FPGA side. And then it is seen that weights of both fully connected layer and output layer are updated in the FPGA as well which means weight values are stored in the FPGA too. It should be stressed that update processes of both momentums and weights are shown separately for the sake of clarity on the image. Each update is done in its own ADAM module in order to reduce loops inside the FPGA.

\subsection{Parallelization}
One of the advantages of FPGA is that its parallelization features are quite abundant and grants the designer very useful tools for acceleration in computing. This is one of the main reasons that an FPGA is used in this design. The training of a deep learning model consists of a huge number of matrix multiplication and addition which causes a long execution time. In order to prevail this problem, several parallelizations could be applied on these matrix operations in order to lessen the long execution times.

This is the main approach of the acceleration process of this design. Briefly, the arrays are partitioned into several blocks and they are calculated in parallel. An example of a parallel matrix multiplication loop in this design can be seen in Listing 1.

\begin{lstlisting}[frame=single, caption=Fully connected layer]
for (i=0 ; i<BATCHSIZE ; ++i){
#pragma HLS UNROLL factor=4
    for (j=0 ; j<LAYERSIZE ; ++j){
#pragma HLS UNROLL factor=4
#pragma HLS PIPELINE
        for (k=0 ; k<POOLMAPLENGTH ; ++k){
            h1[i][j] += v[i][k] * W1[k][j];
        }
    }
}
\end{lstlisting}

It can be seen that the first dimension of the convolution products and the second dimension of the weights of the fully connected layer are read by parts in parallel by unrolling the regarding loops by the factor of 4. On top of this, the multiplication part is calculated in pipelined manner. A more explicit code piece of this calculation can be visualized in Listing 2.

\begin{lstlisting}[frame=single, caption=Explicit code of unrolling fully connected layer]
for (i=0 ; i<BATCHSIZE ; i+=4){
    for (j=0 ; j<LAYERSIZE ; j+=4){
#pragma HLS PIPELINE
        for (k=0 ; k<POOLMAPLENGTH ; k++){
            h1[i][j]     += v[i][k] * W1[k][j];
            h1[i][j+1]   += v[i][k] * W1[k][j+1];
            h1[i][j+2]   += v[i][k] * W1[k][j+2];
            h1[i][j+3]   += v[i][k] * W1[k][j+3];
            
            h1[i+1][j]   += v[i+1][k] * W1[k][j];
            h1[i+1][j+1] += v[i+1][k] * W1[k][j+1];
            h1[i+1][j+2] += v[i+1][k] * W1[k][j+2];
            h1[i+1][j+3] += v[i+1][k] * W1[k][j+3];
            
            h1[i+2][j]   += v[i+2][k] * W1[k][j];
            h1[i+2][j+1] += v[i+2][k] * W1[k][j+1];
            h1[i+2][j+2] += v[i+2][k] * W1[k][j+2];
            h1[i+2][j+3] += v[i+2][k] * W1[k][j+3];
            
            h1[i+3][j]   += v[i+3][k] * W1[k][j];
            h1[i+3][j+1] += v[i+3][k] * W1[k][j+1];
            h1[i+3][j+2] += v[i+3][k] * W1[k][j+2];
            h1[i+3][j+3] += v[i+3][k] * W1[k][j+3];
        }
    }
}
\end{lstlisting}

It can be seen that the arrays have 4 accesses in the same loop body that is parallelized by pipelining.

Arrays on an FPGA are stored in memory blocks that have ports for input and output providing access to the arrays. If this body loop is desired to be computed in one computation cycle, all of the accesses are needed to be open for input and output. This could be done by either increasing ports of the block memories or partition the arrays so that each access is ready in memory at the time of need. Because of the port amount to these block memories is limited and cannot be increased without physical effect, arrays need to be partitioned by the desired access amount. In our case, the arrays need to be parted into 4 in order to get ready for parallel computation. The more the arrays are partitioned, the more parallel they can be accessed and used in calculations.

Another code piece from the design can be seen in Listing 3.

\begin{lstlisting}[frame=single, caption=Output layer]
for (i=0 ; i<BATCHSIZE ; i++){
#pragma HLS UNROLL factor=4
    for (j=0 ; j<CLASSSIZE ; j++){
#pragma HLS UNROLL factor=10    
#pragma HLS pipeline
        for (k=0 ; k<LAYERSIZE ; k++){
            h2[i][j] += h1[i][k] * W2[k][j];
        }
    }
}
\end{lstlisting}

\begin{table*}[htb!]
\centering
\caption{Execution times for one epoch}
\begin{tabular}{|c|c|c|c|c|c|c|c|c|}
\hline
System        & \multicolumn{8}{c|}{\begin{tabular}[c]{@{}c@{}}Execution Time\\ (sec)\end{tabular}}                                                                                                                                                                                                         \\ \hline
              & \multicolumn{4}{c|}{Training}                                                                                                                & \multicolumn{4}{c|}{Test}                                                                                                                    \\ \hline
              & Host & FPGA & \begin{tabular}[c]{@{}c@{}}Total\\ (Sequential)\end{tabular} & \begin{tabular}[c]{@{}c@{}}Total\\ (System Level Pipelined)\end{tabular} & Host & FPGA & \begin{tabular}[c]{@{}c@{}}Total\\ (Sequential)\end{tabular} & \begin{tabular}[c]{@{}c@{}}Total\\ (System Level Pipelined)\end{tabular} \\ \hline
Tensorflow    & 5.3 & -    & 5.3                                                          & -                                                              & 0.8 & -    & 0.8                                                          & -                                                              \\ \hline
This Work & 1.9 & 2.3  & 4.2                                                          & 2.3                                                              & 0.3 & 0.02 & 0.32                                                         & 0.3                                                              \\ \hline
\end{tabular}
\end{table*}

\begin{table*}[htb!]
\centering
\caption{Resource utilizations for FPGA}
\begin{tabular}{|c|c|c|c|c|c|c|c|c|c|c|}
\hline
                                                           & \multicolumn{5}{c|}{Training}          & \multicolumn{5}{c|}{Test}            \\ \hline
                                                           & BRAM  & DSP & FF      & LUT     & URAM & BRAM & DSP & FF     & LUT     & URAM \\ \hline
Amount                                                     & 1,421 & 510 & 514,899 & 186,926 & 56   & 424  & 188 & 81,170 & 126,238 & 0    \\ \hline
\begin{tabular}[c]{@{}c@{}}Utilization\\ (\%)\end{tabular} & 94    & 25  & 92      & 67      & 100  & 28   & 9   & 14     & 45      & 0    \\ \hline
\end{tabular}
\end{table*}

In this example, output layer in the forward pass is calculated in parallel similar to fully connected layer. This is the main parallelization method throughout the whole design. As also it can be seen from Listing 1 and Listing 3, it should be stressed that the dimension of mini-batch size, convolution product and fully connected layer are partitioned into 4 and the dimension of class size is partitioned into 10 which is a complete partition since class size of this problem is 10.

\subsection{Memory Access}
In this subsection, it is aimed to clarify in which way the arrays are partitioned and stored inside FPGA.

As it can be seen from Listing 2, the arrays are accessed in parallel however the indices of the accesses are sequential. Hence, the partitions need to be done so that each index can be accessed at the same time, i.e they need to be at the same level of the partitioned arrays. This is done by the feature of cyclic partition. A basic demonstration of cyclic partition can be seen in Fig. 4.

\begin{figure}[htbp]
\centering
\includegraphics[width=80mm]{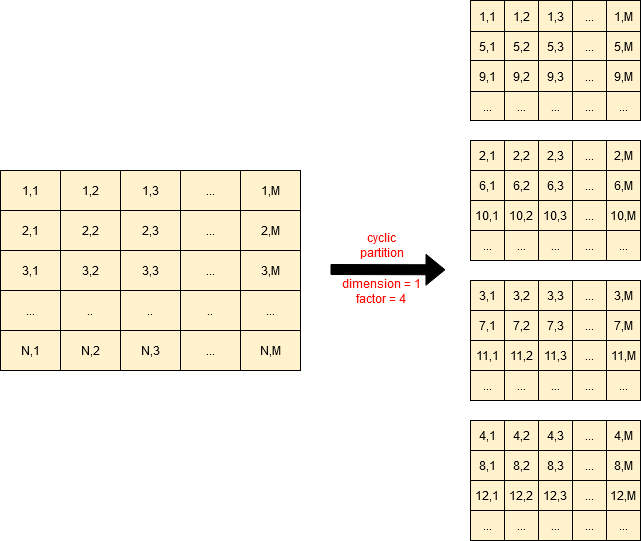}
\caption{Cyclic partition}
\label{fig}
\end{figure}

It can be seen that all partitions have the values of unroll factor of 4 at the same level. The example is solely in dimension 1; nonetheless, it can be applied in dimension 2 as well.

In order to fit into the target FPGA, the partitions are distributed in different types of RAM blocks. The weights are stored in Unified RAM(URAM) pieces. They are quite scarce in the FPGA yet they provide a faster access than regular Block RAMs (BRAM). S\_AXILITE interface reduces the BRAM memory usage for the arrays passed from host to FPGA so BRAM memory usage is used for the other arrays on FPGA. In order to increase throughput while doing so, dual port BRAMs are used. In dual port BRAMs, one port is designated for input and one port is designated for output whereas in single port BRAMs, the only port is used for both input and output.

Finally, in order to limit the LUT register usage to fit into the board, multiplier and adder cores are limited to 25, which may have a decreasing effect on performance, caused almost no material effect on the performance.

\section{Results}
In this section, the execution times for reference with Tensorflow on host computer and this design are compared. Also, the resource utilities for these results are given for this design on FPGA side. It is to be highlighted that the model trained on Tensorflow is identical to the model trained on the design in every aspect including untrained constant kernel in convolution. In addition, the model on Tensorflow is built with the Python version. 64-bit floating point double is used as datatype in the demonstrations. Moreover, the execution times are calculated for just one epoch on the whole dataset. The results can be seen in Table 4.

In the table, two kinds of final execution times can be seen: sequential and system level pipelined. Sequential execution is obtained when each of the parts of the design waits until the other completes its process. In other words, host part applies the convolution, prepares the images and the passes them to FPGA part. It does not prepare the next mini-batch until FPGA is done with its training or inference. Thus their execution times are added on top of each other. System level pipelined version is obtained when the parts are allowed to work in parallel. In this approach, host does not wait until FPGA finishes its process. It continuously produces its convolved mini-batches and sends it to FPGA when it is ready to take the next mini-batch to train or infer. Thus, the execution time of the whole system depends on the slower side of the system. For the training, it completes its whole task faster than FPGA so the latency of the system is the latency of the FPGA. For the testing, the opposite is valid. The latency of the system is the latency of host. 

It can be seen that this design passes Tensorflow on host both in sequential and system level pipelined version. Furthermore, as system level pipelining is valid, the system is approximately 2 times faster than Tensorflow on host.

In Table 5, the resource utilities of FPGA side is given. It can be seen that the resources are utilized in the upper limits in training yet is fits into the chip. The utility usage is remarkably less than training.

This paper leans on acceleration of the training; however, it is not solely enough to accelerate the system. It is expected to give accurate results after the acceleration is done. The accuracy results can be seen in Table 6 for both Tensorflow on host computer and this system.

\begin{table}[htb!]
\centering
\caption{Accuracy}
\begin{tabular}{|c|c|c|c|}
\hline
System        & \multicolumn{3}{c|}{Accuracy (\%)} \\ \hline
              & 1 Epoch   & 5 Epoch   & 10 Epoch   \\ \hline
Tensorflow    & 91        & 93        & 94         \\ \hline
This Work & 90        & 92        & 94         \\ \hline
\end{tabular}
\end{table}

It can be seen that both of the systems converge to the same level of accuracy after 10 epochs. The difference that is seen in early epochs arises from that the weights are randomly initialized in both of the systems. Moreover, the training set is not shuffled between the epochs for both of the systems.

\section{Discussions}
This work shows the potential of FPGA usage in deep learning problems. The results show that the execution times can be drastically lowered without so much accuracy loss in the end. Since acceleration on whole training contains inference, the same hardware piece may accelerate inference as well.

There are improvements that can be done on this work such as acceleration of convolution where convolution kernel is trained too, usage of stream interface and dataflow and different data types such as 16-bit floating point half or fixed point data types. The usage of less precision data types may end up with a decrease in accuracy but increase in performance a lot. In addition, different systems like GPU-FPGA hybrids could be designed in order to improve performance.

\end{document}